\documentclass[10pt,conference]{IEEEtran}

\usepackage[pdftex]{graphicx}

\usepackage[cmex10]{amsmath}
\usepackage{amssymb}
\usepackage[font=footnotesize,caption=false]{subfig}
\usepackage{url}
\usepackage{etoolbox}
\usepackage[super]{nth}
\usepackage{multirow}
\apptocmd{\thebibliography}{\scriptsize}{}{}

\newcommand{\squeezeup}{\vspace{-2.5mm}}

\begin{document}

\title{Convolutional Neural Networks for Font Classification}

\author{\IEEEauthorblockN{Chris Tensmeyer, Daniel Saunders, and Tony Martinez}
\IEEEauthorblockA{Dept. of Computer Science\\
Brigham Young University\\
Provo, USA\\
tensmeyer@byu.edu danielsaunders@byu.edu martinez@cs.byu.edu}
}

\maketitle

\begin{abstract}
Classifying pages or text lines into font categories aids transcription because single font Optical Character Recognition (OCR) is generally more accurate than omni-font OCR.
We present a simple framework based on Convolutional Neural Networks (CNNs), where a CNN is trained to classify small patches of text into predefined font classes.
To classify page or line images, we average the CNN predictions over densely extracted patches.
We show that this method achieves state-of-the-art performance on a challenging dataset of 40 Arabic computer fonts with 98.8\% line level accuracy.
This same method also achieves the highest reported accuracy of 86.6\% in predicting paleographic scribal script classes at the page level on medieval Latin manuscripts.
Finally, we analyze what features are learned by the CNN on Latin manuscripts and find evidence that the CNN is learning both the defining morphological differences between scribal script classes as well as overfitting to class-correlated nuisance factors.
We propose a novel form of data augmentation that improves robustness to text darkness, further increasing classification performance.
\end{abstract}


\section{Introduction}

Deep Convolutional Neural Networks (CNNs) have been successfully applied to many problems in Document Image Analysis.
These areas include whole image classification~\cite{harley15,afzal15,kang14}, image preprocessing~\cite{pastor15}, script identification~\cite{shi16}, and character recognition~\cite{simard03}.
The success of CNNs has been attributed to their ability to learn features in an end-to-end fashion from large quantities of labeled data.

In this work, we present a simple CNN based framework for classifying page images or text lines into font classes.
Handling multiple fonts is a challenge in Optical Character Recognition (OCR), as the OCR system must handle large variations in character appearance due to differences in font.
If text lines are labeled with a font class, then a specialist OCR system for that font can potentially achieve higher recognition rates than an OCR system trained on many fonts~\cite{shi97,baird94}.

In this framework, a CNN is trained to classify small image patches into font classes.
At prediction time, we densely extract patches from the test image and average font predictions over individual patch predictions.
Although this method is simple, we achieve 98.8\% text line accuracy on the King Fahd University Arabic Font Database (KAFD) for 40 type faces in 4 styles and 10 different sizes~\cite{luqman14}.
The best previous result is 96.1\% on a subset of 20 type faces~\cite{luqman14}.
We also demonstrate state-of-the-art performance with 86.6\% accuracy on the Classification of Latin Medival Manuscripts (CLaMM) dataset, where the highest previously reported accuracy is 83.9\%~\cite{cloppet16}.

In addition to showing that CNNs perform well at font classification tasks, we perform an in-depth analysis of the features learned by the CNN.
Though CNNs are black box models, we can gain an understanding of what features are used for classification by measuring output responses as we vary characteristics of the input images.
Such an analysis can demonstrate whether the CNN is overfitting to nuisance factors of the collection of documents it was trained on. 

For example, the CLaMM dataset contains 12 scribal script classes defined by expert paleographers that are handwriting styles that differ in character allographs and morphological shape~\cite{cloppet16}.
However, we find that CNNs trained on CLaMM are sensitive to how dark the text is.
This is undesirable because the CNN may be applied to novel document collections that have a different bias w.r.t. text darkness.
We provide a solution to this problem using a new form of data augmentation, which also improves performance on CLaMM.
We also find that CNNs can be sensitive to other factors such as inter-line spacing and presence of non-textual content.

\section{Related Works}

\label{sec:related_works}

We review the literature for two tasks: font classification and analyzing what features a CNN has learned.
Though the classes in CLaMM are referred to as \emph{script classes}, we feel that this task is more akin to font classification than what is traditionally described as script classification.
Traditional script classification deals with distinguishing different writing systems or character sets (e.g. Chinese vs Latin vs Arabic), while script classes in CLaMM are all Latin script.
However, in keeping with the terminology introduced in~\cite{cloppet16}, we use the term \emph{script} to refer to a class category in CLaMM.
We also use the term \emph{font class} to refer to typefaces (e.g. Arial) rather than combinations of typeface, size, and style (e.g. bold).

\subsection{Font Classification}

Zramdini and Ingold presented a font recognition system based on the statistics of connected components, achieving 97.35\% accuracy over English text lines for 10 font classes~\cite{zramdini98}.
Zhu et al. posed font recognition as texture identification and used Gabor Filters to achieve 99.1\% accuracy on text blocks for both English and Chinese text~\cite{zhu01}.
Fractal Dimension features were introduced in~\cite{moussa10} for Arabic font classification and resulted in 98\% accuracy for 10 font classes.
Luqman et al. used log-Gabor filter features extracted at multiple scales and orientations to obtain 96.1\% accuracy on 20 fonts in the large scale KAFD dataset.

More recently, deep learning techniques based on Convolutional Neural Networks (CNN) and Recurrent Neural Networks (RNN) have been proposed for font classification.
Tao et al. used a combination of CNN and 2D RNN models to classify single Chinese characters into 7 font classes with 97.77\% accuracy~\cite{tao16}.
Pengcheng et al. classified handwritten Chinese characters into 5 calligraphy classes with 95\% accuracy using deep features extracted from a CNN pretrained on natural images~\cite{pengcheng17}.
Classifying calligraphy classes is similar to classifying script types in CLaMM, though CLaMM uses Latin script classes with page-level ground truth.

\subsection{Analysis of Learned Features}

There are a variety of techniques used to examine features learned by CNNs, developed primarily in the context of natural images.
In~\cite{simonyan13}, canonical class images are obtained by gradient descent over input pixels to maximally excite output class neurons.
Similarly, class sensitivity maps are visualized by computing via backpropagation the first partial derivative of the target output class w.r.t. all input pixels.
Zeiler and Fergus proposed a \emph{deconv} visualization approach where hidden representations are projected back into the input space~\cite{zeiler14}.
Additionally, they visualize intermediate neurons by finding the top-N maximally exciting input patches for each neuron.
For quantitative analysis, Zeiler and Fergus measured how CNN outputs change in response to certain transformations of the input image (e.g. rotation, translation)~\cite{zeiler14}.
We perform a similar analysis, specific to the domain of document images, to see how sensitive CNNs are to noise factors in CLaMM.

\section{Methods}
\begin{figure}
\centering
\subfloat[Arabswell]{\fbox{\includegraphics[width=0.13\textwidth]{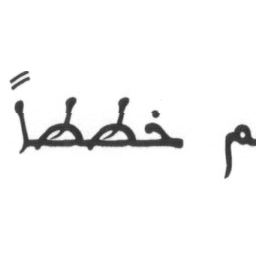}}} \hspace{3pt}
\subfloat[Times New Roman]{\fbox{\includegraphics[width=0.13\textwidth]{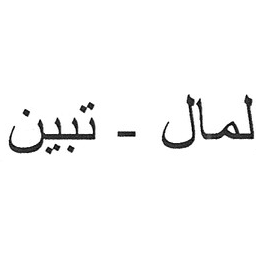}}} \hspace{3pt}
\subfloat[Arial]{\fbox{\includegraphics[width=0.13\textwidth]{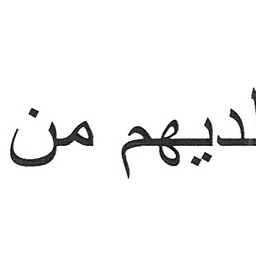}}}

\subfloat[Humanistic]{\includegraphics[width=0.14\textwidth]{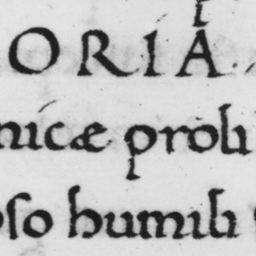}} \hspace{4pt}
\subfloat[Hybrida]{\includegraphics[width=0.14\textwidth]{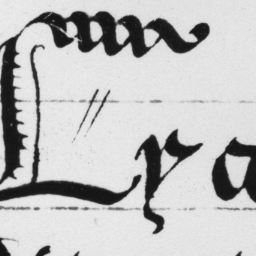}} \hspace{4pt}
\subfloat[Uncial]{\includegraphics[width=0.14\textwidth]{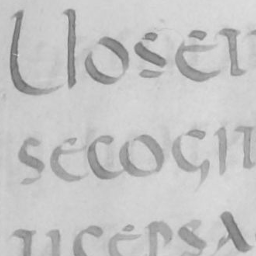}}

\caption{Example 256x256 patches from KAFD (top) and CLaMM (bottom).  Subcaptions refer to the class label of the patch.}
\label{fig:datasets}
\squeezeup
\squeezeup
\end{figure}
\label{sec:methods}

In this work, we classify large document images into script or font classes using CNNs.
Because inputting entire high resolution images to a CNN is computationally slow, requires large GPU memory, and requires more training data, we resort to a patch classification scheme.
We train a CNN to classify individual 227x227 patches into font/script classes.
To obtain a classification for a large test image, we densely extract overlapping 227x227 patches on a regular square grid with 100 pixels between patch centers.
The CNN produces a probability distribution over the classes for each patch, which distributions are uniformly averaged to obtain the final classification.

For training data, we extract 256x256 patches at a stride of 42 pixels from the training images and label patches with the class of the image it was taken from.
During training, a random 227x227 crops from these patches are inputted to the CNN. 
Some training images are set aside as a validation set, which is used to select the best performing model.

In this work, we compare two CNN architectures.
The first is the AlexNet architecture composed of 5 convolution layers followed by 3 fully connected layers~\cite{krizhevsky12}.
Each layer takes as input the output of the previous layer:
\begin{equation}
x_{\ell} = F_{\ell}(x_{\ell - 1}, \theta_{\ell})
\end{equation}
where $x_{\ell}$ is the input to the $\ell^{th}$ layer, $F_{\ell}$ is a function to be learned parameterized by $\theta_{\ell}$.
The $F_{\ell}$ performs either a convolution or a matrix multiplication followed by $\operatorname{ReLU}(z) = \operatorname{max}(z, 0)$, and sometimes pooling and local response normalization operations~\cite{krizhevsky12}.

The other architecture is the state-of-the-art ResNet-50~\cite{he16}, which was used to win the 2015 ImageNet Large Scale Visual Recognition Challenge (ILSVRC).
Layers in a ResNet learn a \emph{residual function}, where $F_{\ell}$ is added to the layer input:
\begin{equation}
\label{eq:res}
x_{\ell} = F_{\ell}(x_{\ell - 1}, \theta_{\ell}) + x_{\ell - 1}
\end{equation}
Residual learning enables deeper networks to be trained with gradient descent optimization as the gradient no longer vanishes or explodes exponentially with layer depth due to the $x_{\ell -1}$ term in Equation~\ref{eq:res}.
For ResNet-50, $F_{\ell}$ is composed of multiple convolutions, ReLU, BatchNorm~\cite{ioffe15}, and sometimes pooling operations.
See~\cite{he16} for the exact model specification.

Additionally, we train CNNs at 7 image scales from 30-100\%, where 100\% is original image size and 50\% is downsampled by a factor of 2..
This is done by downsampling training images before extracting 256x256 patches and by downsampling tests images for classification.
For smaller image scales, more characters are available per 256x256 training patch, but with less detail.
For CLaMM experiments, we ensemble networks by averaging predictions of two CNNs trained from different random initializations.
For training details, all CNNs are trained with Stochastic Gradient Descent (SGD) using 0.9 momentum and L2 weight decay of 0.0005.
The number of iterations was 350K-650K based on architecture and dataset, with mini-batches of size 40-64.

\subsection{Datasets and Evaluation}

We use two datasets in this work, Classification of Latin Medieval Manuscripts (CLaMM)~\cite{cloppet16} and the King Fahd University Arabic Font Database (KAFD).
Example training patches from each dataset are shown in Figure~\ref{fig:datasets}.

The CLaMM dataset was introduced in the recent \emph{ICFHR2016 Competition on the Classification of Medieval Handwritings in Latin Script}~\cite{cloppet16}.
It is comprised of 2000 training images and 1000 test images (for task 1) in grayscale format.
Each image is approximately 1700x1200 pixels (300 dpi) and is composed of handwritten text, background space, and graphics.
The training and test images are annotated with a single \emph{script class}.
There are 12 script classes representing different styles of character shapes used by scribes in producing handwritten manuscripts in Europe in the years 500 C.E. to 1600 C.E.
For CLaMM, we report model accuracy over the 1000 test images.

KAFD is comprised of 115,068 scanned pages of printed Arabic text in grayscale format divided among 40 font classes (e.g. Arial).
Each font class contains pages that differ in font size (8-24 point) and style (regular, bold, italic, bold and italic).
Though KAFD is available in four resolutions, we opted to use only 300dpi images for computational reasons.
We used the designated train/validation/test split and evaluated model accuracy on both the page and line formats of the dataset.
Because line images are less than 256 pixels in height, we pad them with white background to be 256 pixels in height.
When extracting patches, we discard patches with minimum value $> 100$ as these patches likely consist of only background.

\subsection{Pretraining on Synthetic Data}

\label{sec:pretrain}

For tasks with limited data, pretraining networks on similar tasks tends to improve performance~\cite{sharif14}.
Pretraining is performed by initializing network weights to those learned on the pretraining task, except for the classification layer, which is initialized randomly due to the tasks having different classes.

For CLaMM, we experimented with pretraining on a 27-class synthetic font recognition task designed to mimic some character differences between CLaMM classes.
We hope that by pre-conditioning the CNN to examine individual characters, it will more easily discriminate script classes based on their defining morphological differences.
23 font classes were based on the \emph{Liberation Serif} font and 4 more were based on other fonts.
Each class deviates from the basic font by a random combination of the following modifications:
\begin{itemize}
\item Only capital glyphs.
\item Vertical translation of certain glyphs.  Some normally descending characters are shifted to be non-descending, and normally non-descending characters become descenders.
\item Substituting glyphs with characters from CLaMM classes.  For example, using the Uncial \emph{A}, a single compartment Cursiva \emph{a}, or long \emph{s} glyph instead of modern \emph{s}.
\end{itemize}

To create training data for a synthetic font class, we rendered random Latin text and added Gaussian noise to foreground characters.
We then inserted the noised text onto blank background pages taken from real historical documents using image interpolation.
The accuracy on the pretraining task itself ranged from 50-85\% depending on CNN architecture and image scale, indicating that the task is non-trivial and  that CNNs are able to focus on individual characters to make classification decisions.

\section{Results}

\label{sec:results}														


\begin{table}
\centering
\begin{tabular}{|c|c|cc|cc|cc|}
\hline
Train    & \multirow{2}{*}{Model}& \multicolumn{2}{c|}{Lines} & \multicolumn{2}{c|}{Page-Lines} & \multicolumn{2}{c|}{Pages} \\
\cline{3-8}
Data                   &         & Patch & Image & Patch & Image & Patch & Image \\
\hline
\multirow{2}{*}{Lines} & AlexNet & 95.3  & 97.1  & 95.3 & 98.7  & 38.5  & 90.8  \\ 
                       & ResNet  & \textbf{97.9} & \textbf{98.8} & \textbf{97.9} & \textbf{99.2} & 58.0 & 92.7 \\ 
\hline
\multirow{2}{*}{Pages} & AlexNet & 86.2  & 88.4  & 86.0  & 91.1  & 58.2 & 98.2 \\ 
                       & ResNet  & 91.9  & 93.2  & 91.7  & 94.8  & \textbf{60.5} & \textbf{98.5} \\ 
		
\hline

\end{tabular}												
\setlength\tabcolsep{6pt}
\caption{Accuracy of CNNs on KAFD test data at both the patch and whole image level.  }
\label{tab:kafd}
\squeezeup
\squeezeup
\end{table}

\begin{figure}
\centering
\subfloat[Segore UI]{\includegraphics[width=.45\textwidth]{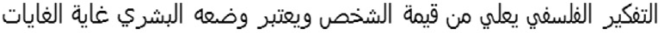}}

\subfloat[Times New Romans]{\includegraphics[width=.45\textwidth]{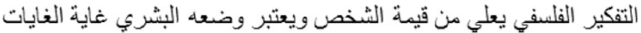}}

\subfloat[Arial]{\includegraphics[width=.45\textwidth]{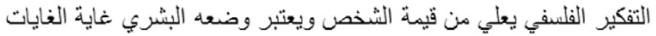}}

\subfloat[Arabic Transparent]{\includegraphics[width=.45\textwidth]{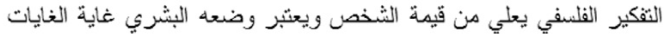}}
\caption{KAFD classes that are easily confused, accounting for approximately 70\% of misclassifications in all trained models.}
\label{fig:kafd_conf}
\squeezeup
\squeezeup
\end{figure}

\begin{figure*}
\centering
\subfloat[]{\includegraphics[width=.3\textwidth]{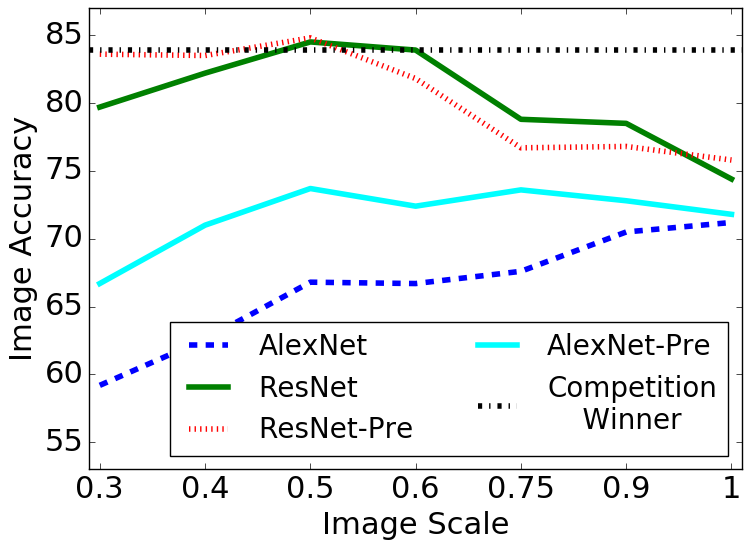}}\hspace{5pt}
\subfloat[]{\includegraphics[width=.3\textwidth]{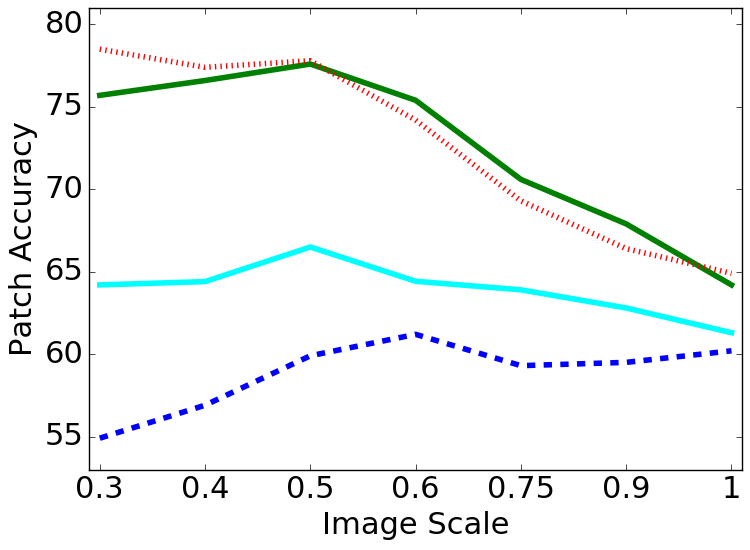}}\hspace{5pt}
\subfloat[]{\includegraphics[width=.3\textwidth]{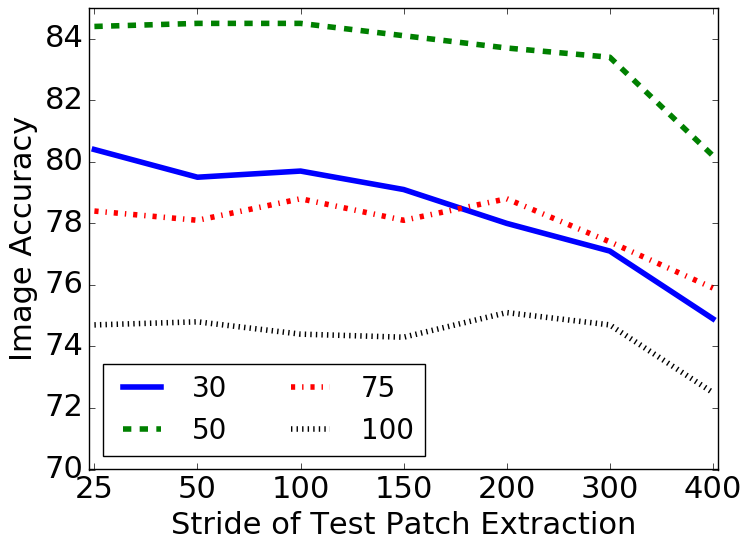}}
\caption{Performance on CLaMM for CNNs trained at difference scales on (a) whole image classification and (b) per-patch classification.  \emph{-Pre} indicates that CNNs were first pretrained on synthetic data (Section~\ref{sec:pretrain}).  A pretrained ResNet trained at 50\% image scale performs best and outperforms the previous best result on this task.  Subfigure (c) shows the effect of patch extraction stride at test time.  Smaller strides are more important for smaller image scales.}
\label{fig:clamm_graphs}
\squeezeup
\end{figure*}

\subsection{KAFD}

On KAFD, we trained 2 ResNets and 2 AlexNet CNNs each on line images and on page images at 100\% image scale.
For each CNN architecture and type of training data, we selected one model using the provided validation data.
Model accuracies for both patches and images are shown in Table~\ref{tab:kafd}.
The \emph{Page-Lines} column shows page level accuracy obtained by averaging predictions over the pre-segmented line images for each page.
For the \emph{Pages} column, test patches are densely extracted from the page image and may contain badly cropped text, leading to lower patch accuracy for this column.

For all types of data, the more powerful Resnet architectures outperforms AlexNet, providing 19-58\% error reduction.
Training and testing on segmented line images leads to the best page level classification at 99.2\% accuracy for ResNet.
When training and testing on densely cropped patches, the best accuracy is 98.5\%, showing that using segmented data leads to a 47\% reduction in error over densely cropped patches.
The best accuracy achieved over line images is 98.8\%.
 
To our knowledge, there are no previously published results on all 40 fonts.
On subsets of 10 and 20 fonts, Luqman et al. achieved 99.5\% and 96.1\% accuracy respectively on line images using log-Gabor features extracted at multiple scales and orientations~\cite{luqman14}.
On the same 20 fonts, a single ResNet model achieves 99.8\% accuracy on line images.

The vast majority of misclassifications on all 40 fonts are made by confusing \emph{Times New Roman} with either \emph{Segore UI}, \emph{Arial}, or \emph{Arabic Transparent}.
These classes account for at least 70\% of errors for patches, line images, and page images.
Figure~\ref{fig:kafd_conf} shows example text for these 4 easily confused classes.
The fonts \emph{Times New Roman}, \emph{Arial}, and \emph{Arabic Transparent} are very similar, and could be grouped together under a single OCR system.
However, \emph{Segore UI} is visually distinctive (e.g. larger holes inside characters) and shows that there is room for improvement in the model.

\subsection{CLaMM}

For CLaMM, Figure~\ref{fig:clamm_graphs} shows the performance of ensembles composed of two models, where each ensemble is trained at a single image scale.
On this task, we see the importance of using the more powerful ResNet architecture.
In general, mid-range image scales perform best, likely because they balance the trade-off between amount of text on each patch and resolution of the text.
Notably, using 50\% image scale with the ResNet architecture yields 84.5\% accuracy, which outperforms the highest reported result of 83.9\% achieved by the ICFHR 2016 CLaMM competition winner~\cite{cloppet16}.

Pretraining on synthetic data significantly improves performance for AlexNet models and shows that this is a viable approach to improving CNNs based on domain knowledge.
For ResNet, pretraining helps for smaller image scales, though it causes a slight decrease in performance for some larger scales.
This is likely because overfitting training data is a bigger problem with smaller scales due to having fewer training patches.
Pretraining on the synthetic data at 50\% image scale leads to an increase of 0.3\% absolute accuracy to reach 84.8\% on this task.
We note that we created only one set of pretraining data and did not tweak the pretraining classes to improve results, either on test or validation data.

We also analyzed the stride at which patches are extracted from test images (Figure~\ref{fig:clamm_graphs}c).
Larger strides require less computation because fewer patches are evaluated, but also result in lower accuracy.
Smaller strides seems to be more important for smaller image scales, likely because fewer patches are extracted from smaller images at any given stride.

\section{Analyzing Learned Features}
\label{sec:features}

In this section, we analyze how sensitive a ResNet model is to noise factors present in the CLaMM dataset.
We take two approaches.
In the first approach, we modify the training data and evaluate the performance of the ResNet on the task images.
In the second, we take the trained ResNet and measure how changing certain characteristics of input patches affects the classification decision for the patch.

\subsection{Modified Training Data}

\begin{figure}
\centering


\subfloat[Background]{\includegraphics[width=0.145\textwidth]{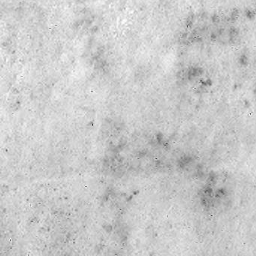}} \hspace{4pt}
\subfloat[Figure]{\includegraphics[width=0.145\textwidth]{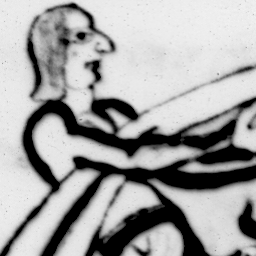}} \hspace{4pt}
\subfloat[Annotation]{\includegraphics[width=0.145\textwidth]{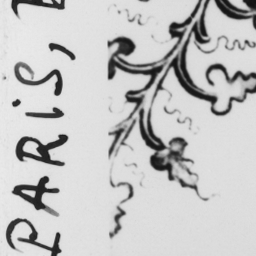}} \hspace{4pt}

\caption{Example of non-textual patches from CLaMM dataset.  Though they do not contain text of the target script class, some of these patches are discriminative of the script class of the page.}
\squeezeup
\squeezeup
\label{fig:noise}
\end{figure}

\begin{table*}
\scriptsize
\centering
\setlength\tabcolsep{3pt}
\begin{tabular}{|l|c|cccccccccccc|}
\hline
                 & Average & Caroline & Cursiva & Half-Uncial & Humanistic & Humanistic & Hybrida & Praegothica & Semihybrida & Semitextualis & Southern  & Textualis & Uncial  \\
Training Set     &         &          &         &             &            & Cursive    &         &             &             &               & Textualis &           &         \\
\hline
CLaMM            & 74.4    & 97.7     & 65.1    & 84.4        & 89.0       & 91.1       & 37.5    & 94.0        & 37.3        & 82.4          & 64.6      & 51.8      & 100      \\
CLaMM-Filtered   & 77.4    & 97.7     & 65.1    & 91.1        & 92.7       & 92.4       & 34.1    & 94.0        & 49.4        & 88.2          & 68.3      & 58.8      & 100      \\
CLaMM-Extended   & 77.7    & 96.5     & 58.1    & 95.6        & 91.5       & 93.7       & 31.8    & 91.7        & 55.4        & 85.3          & 79.3      & 56.5      & 100      \\
CLaMM-Noise      & 21.5    & 15.1     & 9.3     & 15.6        & 39.0       & 26.6       & 18.2    & 48.8        & 15.7        & 22.1          & 7.3       & 20.0      & 21.8     \\
\hline
\end{tabular}
\setlength\tabcolsep{6pt}
\caption{Whole image class accuracies on CLaMM test images with ResNet at 100\% image scale for modified training sets.}
\label{tab:modified_training}
\squeezeup
\squeezeup
\squeezeup
\end{table*}

Because training and testing patches are densely extracted, not all patches contain text that can be used to discriminate the page image's script class.
Such patches may contain only background, figures, or institutional annotations (see Figure~\ref{fig:noise}).
To measure whether these patches postively or negatively affect model performance, we manually annotated\footnote{These annotations are available for download at \url{http://axon.cs.byu.edu/clamm}} each of the 2000 training images using the PixLabler Tool~\cite{saund09} and foreground masks computed using Otsu binarization~\cite{otsu75}.

We created three modified training sets with the annotations:
\begin{enumerate}
\item CLaMM-Filtered (12 classes) - Non-textual patches are removed from the training set.
\item CLaMM-Extended (15 classes) - Non-textual patches are reassigned to one of \emph{background}, \emph{figure}, \emph{annotation}.  At test time, predictions for these classes are not allowed
\item CLaMM-Noise (13 classes) - All textual patches are removed from the 12 script classes (manually verified), leaving only non-textual patches as examples of the script classes.  An additional \emph{Text} class is composed of textual patches drawn from all classes.  At test time, predictions for the \emph{Text} class are not allowed.
\end{enumerate}

We trained an ensemble of two ResNets at 100\% image scale (test patch stride of 100) on the three modified training sets and on the unmodified training set.
We chose 100\% image scale so that \emph{CLaMM-Noise} could have a sufficient number of training patches.
The per-class and average accuracies are shown in Table~\ref{tab:modified_training}.

Interestingly, we see that either filtering or relabling non-textual patches increases accuracy by 3\%, indicating that it is detrimental to label non-textual patches as examples of script classes.
Because non-textual patches are distinctly labeled in \emph{CLaMM-Extended}, the ResNet can minimize their impact at test time.
This data preprocessing outperforms that of \emph{CLaMM-Filtered}, where the ResNet at test time must classify non-textual patches into one of the 12 script classes.
The results on \emph{CLaMM-Noise} demonstrate that some script classes can be discriminated based on non-textual content, as the average accuracy is 21.5\%, compared to 8.3\% for random predictions.
For example, images containing \emph{Praegothica} and \emph{Humanistic} scripts frequently have large decorated figures, which can be sufficient to classify the images.
Some other reasons for performance above random chance include correlation of figures, background intensity, presence of noise (e.g. bleed through) with page level script classes.
This shows that CNNs have the potential to overfit to particular characteristics of the collection used to train (and evaluate) the model, so caution should be exercised when applying models to novel collections.

\subsection{Text Darkness}

We also examined learned features by varying input patches and measuring per-patch output accuracy.
In preliminary analysis, we found two nuisance factors that influence classification decisions: text darkness and inter-line spacing.

\begin{figure}
\centering
\includegraphics[width=0.45\textwidth]{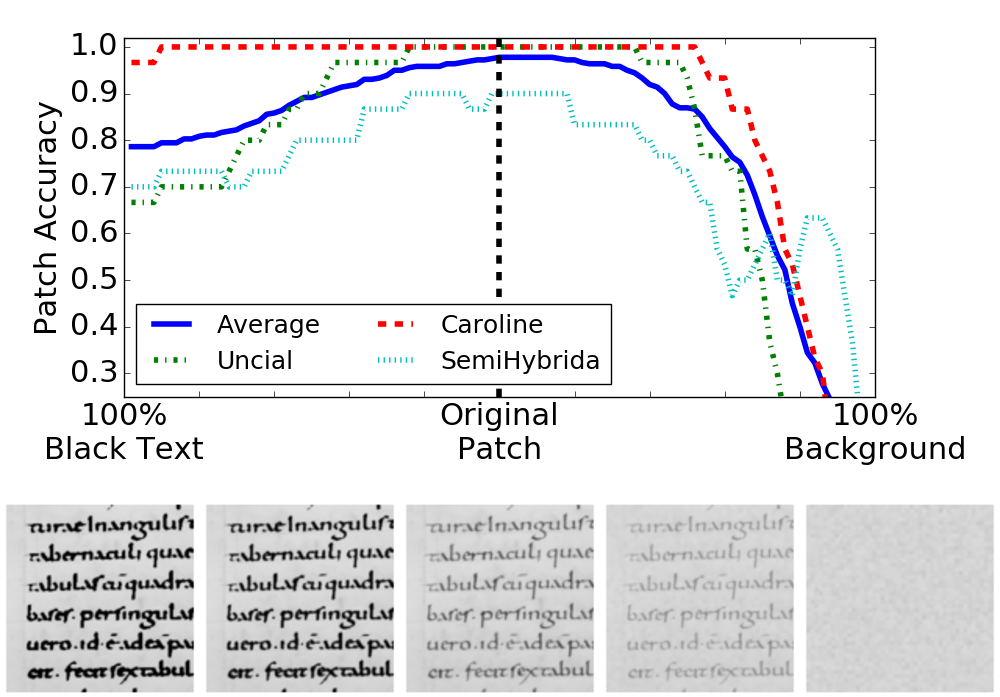}
\caption{Patch Accuracy as a function of text darkness.  Patches with varying text darkness below are positioned relative to the x-axis of the graph.  Some classes are more or less sensitive to these effects.}
\label{fig:darkness}
\squeezeup
\squeezeup
\end{figure}

\begin{figure}
\centering
\includegraphics[width=0.4\textwidth]{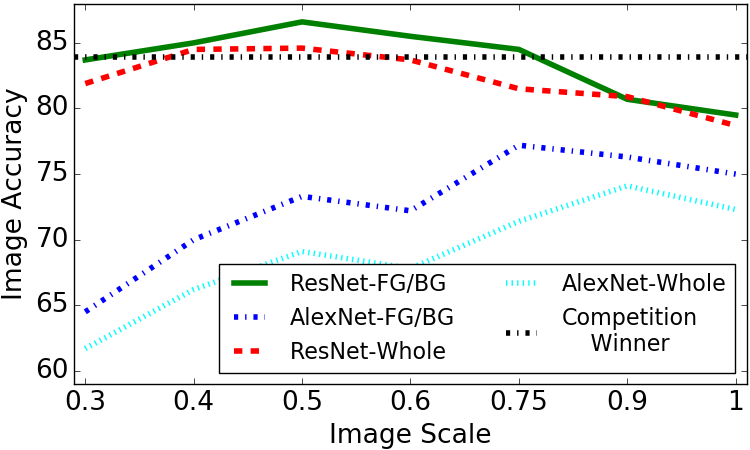}
\caption{Comparing independently varying background and foreground intensity (\emph{-FG/BG}) vs varying whole image intensity (\emph{-Whole}) on CLaMM.}
\label{fig:jitter}
\squeezeup
\squeezeup
\end{figure}

To test the effect of text darkness on patch-accuracy, we extracted 30 patches of text from each class and computed reasonable foreground masks using Otsu binarization~\cite{otsu75}.
For each patch, we produced a series of 100 patches, where 50 have darker text and 50 have lighter text compared to the original patch.
To make text darker, we subtracted constant values from all foreground pixels and clipped values at 0.
The value of the constant was linearly varied such that the intensity of all foreground pixels in the darkest image are uniformly 0.
To make the text lighter, we produced linear combinations of the original patch and an estimated background image constructed by averaging together sampled background patches.
These linear combinations range from 100\% original image to 100\% background estimate.
We chose interpolation over directly lightening text to avoid making any foreground pixels lighter than the surrounding background.

We ordered these 100 modified patches with the darkest as the \nth{1}, the original patch as the \nth{50}, and the lightest patch as the \nth{100}.
We classified each modified patch and recorded the average accuracy for each darkness/lightness level for all patches and each class (Figure~\ref{fig:darkness}).
Overall, classification is robust to small changes, but there are sharp decreases for larger changes.
For example, the training data for \emph{Uncial} script has uniformly light text, with an average foreground intensity of 130 (other classes have average intensity $\sim$50).
We observe that when \emph{Uncial} text is darkened or lightened, more errors are made because such examples are not present in the training data.
However, lightness or darkness of foreground text is not a defining characteristic of the script, but is an artifact of writing instrument, ink composition, and document preservation conditions.
The majority of other scripts, such as \emph{SemiHybrida}, exhibit similar sensitivities, while \emph{Caroline} script appears robust to both darkening and lightening of text.

To make CNNs robust to varying text darkness, we apply a novel form of data augmentation.
While it is common to randomly brighten or darknen input images on-the-fly during training (e.g.~\cite{krizhevsky12}), we independently lighten or darken foreground and background pixels based on masks computed using Otsu binarization~\cite{otsu75}.
Specifically, each time an image is input to the CNN, we choose either foreground or background with equal probability.
Then, we draw a random value from a Gaussian distribution with $\mu=0, \sigma=30$ and add that value to the grayscale pixel values of the selected region.

In Figure~\ref{fig:jitter} we compare this scheme to simply brightening or darkening the whole image.
In general, independently varying foreground and background leads to significant performance gains for all image scales.
In particular, we reach a new record at 86.6\% accuracy on CLaMM, which exceeds the previous state-of-the-art by an absolute 2.7\%.

\subsection{Line Spacing}

\begin{figure}
\centering
\includegraphics[width=0.23\textwidth]{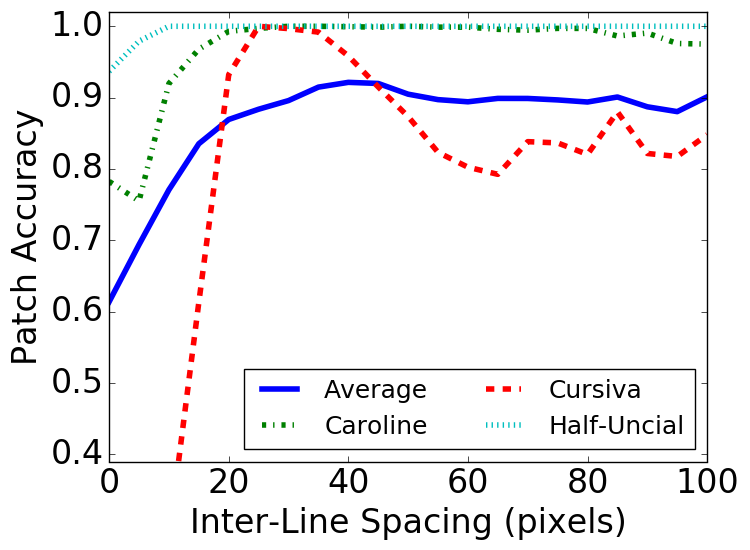}
\includegraphics[width=0.23\textwidth]{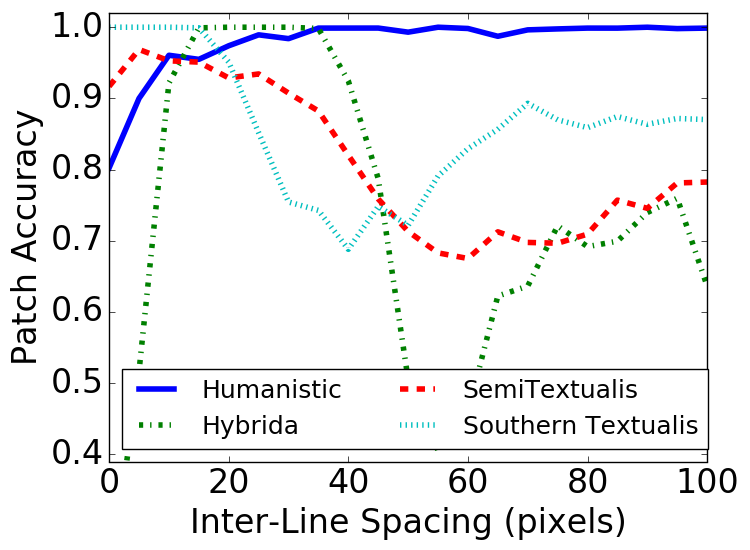}
\caption{Patch accuracy as a function of text line spacing for selected images.  Split into two graphs for clarity.}
\label{fig:line_spacing}
\squeezeup
\squeezeup
\end{figure}

We also found that CNNs are sensitive to the inter-line spacing of text for certain classes.
For each class, we manually extracted text lines and backgrounds from two images and were able to render those text lines at various spacings on the original backgrounds.
We then measured patch-accuracy as a function of text line spacing for each image.

Patch accuracies of selected examples of sensitive images are shown in Figure~\ref{fig:line_spacing}.
The majority of images examined are most easily classified at line spacing of 20 pixels.
For 4 classes, accuracies stayed above 95\% for line spacings greater than 20 pixels.
For 3 of these classes, sharp drops in accuracy were observed for spacings less than 20 pixels (especially at 0 pixels).
Only \emph{Praegothica} images were completely invariant to line spacing.
Trends are not necessarily tied to the script class, as we observed that for 5 of 12 classes, the two images examined had drastically different sensitivities to line spacing.

We note that inter-line spacing is not part of the morphological definition of CLaMM classes.
Therefore it would be desirable to have predictive models that are not sensitive to line spacings.
Though we have not experimentally verified such, we hypothesize that data augmentation where line spacings are stochastically altered (e.g. with seam carving) would give CNNs more invariance to this nuisance factor and potentially make them more accurate for application to novel collections.

Additional experiments (omitted for space) suggested CNNs are sensitive to the line height (i.e. font size) of text lines.

\section{Conclusion}
\label{sec:conclusion}

We have presented a simple patch based classification framework for line image and page image font classification.
We have shown that the ResNet architecture in our framework gives state-of-the-art performance by exceeding previously published results in Arabic font classification on the KAFD dataset and in Latin scribal script classification on the CLaMM dataset.
We performed an analysis of the sensitivities of ResNet to nuisance factors in the CLaMM dataset, such as non-textual patches, text darkness, and line spacing.
In the case of text darkness, we proposed novel data augmentation based on independently varying foreground and background intensities, which leads to improved model robustness and performance.

{\scriptsize
\newcommand{\BIBdecl}{\setlength{\itemsep}{0.25 em}}
\bibliographystyle{IEEEtran}
\bibliography{bib}

}
	
\end{document}